\documentclass{article} 
\usepackage{nips12submit_e}
\usepackage{times}
\usepackage{latexsym,amssymb,amsmath,multicol}
\usepackage{caption}
\usepackage{subcaption}
\usepackage[urlcolor=blue,bookmarks=false,pdfpagemode=None]{hyperref}

\usepackage[pdftex]{graphicx}
\usepackage{epstopdf}
\usepackage[show]{ed}
\usepackage{tikz}
\usetikzlibrary{positioning}
\usetikzlibrary{fit}

\graphicspath{{fig/}}
\renewcommand{\vec}[1]{\mathbf{#1}}
\def\({\left(}
\def\){\right)}

\def\beq{\begin{equation}}
\def\eeq{\end{equation}}

\def\ds{\displaystyle}

\def\ds{\displaystyle}
\def\[{\left[}
\def\]{\right]}
\def\({\left(}
\def\){\right)}

\def\be{\begin{enumerate}}
\def\ee{\end{enumerate}}

\def\ba{\begin{align}}
\def\ea{\end{align}}

\title{Regularization and nonlinearities for neural language models: when are they needed?}

\author{
Marius Pachitariu \\
Gatsby Computational Neuroscience Unit \\
University College London, UK\\
\texttt{marius@gatsby.ucl.ac.uk} \\
\And
Maneesh Sahani \\
Gatsby Computational Neuroscience Unit \\
University College London, UK\\
\texttt{maneesh@gatsby.ucl.ac.uk} \\
}

\nipsfinalcopy 

\begin{document}

\maketitle

\begin{abstract}
     Neural language models (LMs) based on recurrent neural networks (RNN) are some of the most successful word and character-level LMs. Why do they work so well, in particular better than linear neural LMs? Possible explanations are that RNNs have an implicitly better regularization or that RNNs have a higher capacity for storing patterns due to their nonlinearities or both. Here we argue for the first explanation in the limit of little training data and the second explanation for large amounts of text data. We show state-of-the-art performance on the popular and small Penn dataset when RNN LMs are regularized with random dropout. Nonetheless, we show even better performance from a simplified, much less expressive linear RNN model without off-diagonal entries in the recurrent matrix. We call this model an impulse-response LM (IRLM). Using random dropout, column normalization and annealed learning rates, IRLMs develop neurons that keep a memory of up to 50 words in the past and achieve a perplexity of 102.5 on the Penn dataset. On two large datasets however, the same regularization methods are unsuccessful for both models and the RNN's expressivity allows it to overtake the IRLM by 10 and 20 percent perplexity, respectively. Despite the perplexity gap, IRLMs still outperform RNNs on the Microsoft Research Sentence Completion (MRSC) task. We develop a slightly modified IRLM that separates long-context units (LCUs) from short-context units and show that the LCUs alone achieve a state-of-the-art performance on the MRSC task of 60.8\%. Our analysis indicates that a fruitful direction of research for neural LMs lies in developing more accessible internal representations, and suggests an optimization regime of very high momentum terms for effectively training such models.
\end{abstract}

\section{Introduction}
The main paradigm shift in language modelling more than 20 years ago moved the field from rule-based systems to statistical, learned models. The most popular such models are based on frequency statistics of short sequences of words, called n-grams. Various smoothing techniques were proposed to solve the fundamental problem of language modelling: most words and combinations of words appear very rarely in language. In fact, from the point of view of n-gram techniques, language modelling is fundamentally a smoothing or in other words a regularization problem. More recently, models based on neural networks (NNLMs) have been shown to result in better representations of language due to their lower dimensional parametrization and higher ability to generalize \cite{Bengio2003}. Recurrent NNLMs are a subclass of NNLMs that achieve even better performance with a clever parametrization of the predictive likelihood function through a recurrent neural network.

However, to yield top perplexity scores, the neural network language models currently still need to be combined with N-gram based models, caching techniques and averaged over an ensemble of different models as shown in \cite{Mikolov2011a}. Furthermore, the RNN based LMs already require very long training times individually and can be slow at run time if the average over an ensemble needs to be computed. Here we show that random dropout based regularization \cite{Hinton2012} improves the performance of RNN LMs on small datasets. Furthermore, a simpler model that we study here, the impulse-response LM (IRLM), achieves equal performance to the nonlinear RNN when both are regularized in this manner. The IRLM is similar to the log-bilinear language model \cite{MnihHinton2007} (LBL) and can also be seen as a learned Long Short Term Memory (LSTM) network \cite{HochSchmid1997}. Special units in the LSTM have a connection strength of 1 to themselves and no connections to the rest of the network, which alleviates the problem of decaying gradients in learning RNNs. Our IRLM is composed exclusively of these special LSTM units, with the generalization that the self-connection strength can be anything between -1 and 1 and it is learned together with the other parameters of the IRLM. In practice, the IRLM learns to incorporate information from very large contexts of up to 50 words, while also capturing local information.
\vspace{-.2cm}
\section{Recurrent neural networks}
\vspace{-.2cm}
\subsection{Backpropagaton through time}
Standard recurrent neural networks are functions of input data. We consider sequential data that comes in the form of discrete tokens, such that the tokens may for example be either characters or words. An RNN function takes the following form:
\begin{align*}
\vec x_t &= f\(\vec W \enspace  \vec y_t + \vec R\enspace  \vec x_{t-1}\),
\end{align*}
where $\vec y_t$ is the data and $\vec x_t$ is the representation computed by the RNN. In the case of language modelling, $\vec y_t$ is a one-hot encoding of the token in position $t$, meaning $\vec y_t$ is a vector of mostly zeros with just a one in the position of the active token. We call $\vec W$ the encoding matrix and $\vec R$ the recurrent or transformation matrix.  $f$ is typically taken to be a sigmoidal nonlinearity such as the logistic $\sigma(x) = 1/(1+\exp(-x))$. It is generally believed that such strong nonlinearities are necessary to model the apparently complicated dependencies in real-world data. The disadvantages of sigmoidal nonlinearities will become apparent when we consider the optimization problem below.
To make a statistical language model out of an RNN, we define a soft-max probability over the tokens in the sequence, such that this probability depends only on the representation $\vec x_t$ computed by the RNN:
\begin{align}
\text{P} \(\vec y_{t+1} | \vec y_t, \vec y_{t-1}, ...\) \propto \exp(\vec y_{t+1}^T \vec Z \vec x_t) \label{eq:PYX},
\end{align}
where $\vec Z$ is a decoding matrix. We use the terminology of encoding, transformation and decoding matrices for $\vec W, \vec R$ and $\vec Z$ due to the similarity with methods based on autoencoders, RBMs or sparse coding for static data. To obtain the likelihood of a full sequence we multiply together the conditional probabilities defined by equation \ref{eq:PYX} for every index $T$ from 1 to the length of the sequence.

The likelihood of the RNN LM can be optimized by gradient descent. Notice that in order to learn $\vec W$ and $\vec R$ the gradient has to be backpropagated through time (BPTT) over successive activations $\vec x_t$. The intermediate gradients at $\vec x_t$ which we call $\vec D_t$ can be computed incrementally in the reverse-time direction
\begin{align}
\vec D_t &= f'(\vec x_t) \circ \(\vec Z^T \enspace  \frac{\partial \text{P}\(\vec y_{t+1}|\vec x_t\)}{\partial \vec y_{t+1}} + \vec R^T \enspace  \vec D_{t+1}\),
\label{eq:BPTT}
\end{align}
where ``$\circ$'' is elementwise multiplication and $f'$ is the derivative. This backward pass for BPTT has a similar functional form to the forward pass of equation \ref{eq:PYX} and the same computational complexity. It has been observed early on in \cite{Bengio1994} that during training the contribution to the gradient of $W$ and $R$ from future times tends to vanish as it is backpropagated through several time steps. This can be understood by considering the Jacobians $\ds \partial \vec x_{t} / \partial \vec x_{t - \tau}$ of the transformations which the RNN is computing sequentially, and noticing that two main effects alter the size of the gradient/Jacobian. First, the transformation matrix itself generally has eigenvalues less than 1 in absolute value in order to be stable. Consequently, the projection of $\vec x^T$ on the eigenvectors of $\vec R$ decays exponentially over time steps, both in the forward direction when computing the likelihood and in the backward direction when computing the gradients, because $\vec R$ and $\vec R^T$ have the same eigenvalues. The second effect comes from the  nonlinearity which further multiplies the gradient $\vec D_t$ elementwise by the gradient of $f$. Typically during learning the sigmoid saturates either at 0 or 1 where the derivative $f'$ is 0, which drastically reduces the contribution to the gradient from future time steps. We hypothesized that this second effect has a much greater influence on BPTT than the eigenvalues of the transformation matrix $\vec R$. In fact, with linear RNNs where $f$ is just the identity, we find that the network easily learns matrices $\vec R$ with 20\% of their eigenvalues larger than 0.9.
\vspace{-.2cm}
\subsection{Optimizing RNNs on character-level language modelling}
\vspace{-.2cm}
We developed the RNN model used here on character-level language modelling where instead of using the RNN to predict words sequentially we use it to predict characters. RNNs can be difficult to train and require many passes through the training data so it is important to have good optimization techniques. In this section we show that training RNNs with very large  values of momentum can optimize the difficult cost function associated with character-level language modelling. Such cost functions were previously shown to be hard to optimize by \cite{Sutskever2011} and \cite{Mikolov2012}. The authors of those studies propose instead a Hessian-Free training method which uses second order information but can be computationally very demanding. The implementation of \cite{Sutskever2011} takes five days to train in parallel on eight high-end GPUs. Instead we push the momentum term to very high values, such that every gradient update is an effective average over the past approximately one million tokens. We also take advantage of GPUs to run many gradient updates and find that the RNN optimizes to prediction levels comparable to those reported with Hessian-Free optimization in \cite{Mikolov2012} on the `text8' dataset, in less than a day on a single high-end GPU. 

We use a version of RNN in which the nonlinearity $f$ has been replaced with a rectifier function of the form $f(x) = \text{max}(0, x)$. Unlike the sigmoidal nonlinearity, rectifier functions have derivative 1 for positive input which means they fully propagate the gradient in equation \ref{eq:BPTT}. This architecture is also currently and independently investigated by \cite{Pascanu2012}, though we were not aware of their results when we ran our own experiments. We found that the standard rectifier nonlinearity they use there is relatively unstable during learning on character-level problems. Furthermore, the results reported in \cite{Pascanu2012} for this architecture are well behind the state-of-the-art character level results reported in \cite{Mikolov2012} with the Hessian-Free optimized M-RNN of \cite{Sutskever2011} (see table \ref{tab:PennChar}). We adopted instead a smoothed version of the rectifier nonlinearity which is differentiable everywhere and still 0 when the input is negative $f(x) = \text{max}(0, x - a \tanh x/a)$ and found that this simple smooth nonlinearity can be used stably with large learning rates. When $a$ is made small this function approaches the standard rectifier nonlinearity. The H-RNN can model highly nonlinear distributions of sequences as shown by its performance on the character-level modelling task. We also used the RNN in word-level experiments. There we reverted back to the standard rectifier nonlinearity, as we did not observe the same instabilities.

To our knowledge this is the first published result showing that stochastic gradient descent (SGD) learning of nonlinear RNNs can achieve the same performance as the second order optimization methods proposed by \cite{Sutskever2011} and evaluated on the text8 dataset by \cite{Mikolov2012}. The training cost of RNN with gradient descent is about 15h on a single high-end GPU card, while the HF-MRNN takes five days on 8 high-end GPUs, so we observe a more than 50 fold speedup for the same predictive likelihoods. Notice that while a very similar RNN architecture is evaluated in \cite{Pascanu2012}, their results are worse than ours, owing perhaps to the small size of the network they use (512 neurons as opposed to our 2048) and the non-smooth version of the rectifier nonlinearity used there. We did not find it necessary to clip the gradients as done in \cite{Pascanu2012}, because the very high momentum term smoothed the gradient sufficiently. Although we did observe sharp falls of the cost function a few times during training, these were not associated with large changes in the parameters, and the network recovered with a few parameter updates to its previous value of the cost function. Large momentum seemed however to be crucial for fast learning. Without it, the network diverged even with much smaller learning rates. The analysis of \cite{Sutskever2013} also suggests momentum as an important technique for training RNNs and deep neural networks, although the authors suggests it should be used in conjunction with a special initialization of the RNN as an Echo State Network \cite{Jaeger2004} but we did not investigate such an initialization. While we did initialize the parameters of the encoder and decoder matrices to relatively large values, we initialized the recurrent weights close to 0. 

In separate experiments we trained the same RNN on the raw version of the same first 100 MB of Wikipedia. We found that SGD training diverged much more easily in that case and required clipping the gradients as is also done in \cite{Pascanu2012}. On closer inspection, we found that all such cases of divergence were associated with very rare events not related to language at all, such as multiple repeats of the same low-probability character like dashes and exclamation marks. With a minimum of preprocessing we removed such events and the RNN was able to learn successfully from raw Wikipedia without clipping gradients in SGD. More research is needed to clarify the actual geometry of neural networks because as \cite{Pascanu2012} points out, the classical image of long narrow ravines might be quite wrong.

As a curious fact, we note that the RNN models we train discover an ingenious solution for maintaining stability during learning. This is further detailed in the supplemental material.
\vspace{-.1in}
\begin{table}
  \centering
    \footnotesize
    \centering
    \begin{subtable}[]{.45\textwidth}
    \centering
    \begin{tabular}{ | l | c |}
        \hline
         & Bits per character \\\hline
         HF-MRNN$^1$  & 1.54 \\ \hline
         RNN$^2$ & 1.80 \\ \hline
         IRLM w/ NN output & 2.12 \\ \hline
         RNN$^3$ & 1.55 \\ \hline
         skipping RNN & \textbf{1.48} \\ \hline
         subword HF-RNN$^1$ & \textbf{1.49}  \\ \hline
    \end{tabular}
    \captionsetup{width=.9 \textwidth}
    \subcaption{\label{tab:PennChar} Character-level models. \\$^1$ as reported in \cite{Mikolov2012}. $^2$ as reported in \cite{Pascanu2012}. $^3$ our implementation.}
    \end{subtable}
    \begin{subtable}[]{.45\textwidth}
    \centering
    \begin{tabular}{ | l | c |}
        \hline
         & Perplexity \\\hline
         KN5  & 298 \\ \hline
         KN5+cache & 228 \\ \hline
         IRLM & 197 \\ \hline
         RNN & \textbf{179} \\ \hline
    \end{tabular}\\
    \subcaption{\label{tab:PennWord} Word-level results on `text8'}
    \end{subtable}   
    \begin{subtable}[]{.45\textwidth}    
    \centering
    \begin{tabular}{ | l | c | c |}
        \hline
         & Perplexity & MRSC  \\ \hline
         KN4  &  & 39\%  		\\ \hline
         LBL &  & \textbf{54.7} \%		\\ \hline
         IRLM & 103.5 & 52.6\% 	\\ \hline
         RNN & \textbf{79} & 51.3\% 	\\\hline
    \end{tabular}
    \subcaption{\label{tab:perpGUT} Word-level results on project Gutenberg}
    \end{subtable}    
    \begin{subtable}[]{.45\textwidth}    
    \centering
    \begin{tabular}{ | l | c | c |}
        \hline
         &  MRSC  \\ \hline
         IRLM  &  54.8\%  		\\ \hline
         RNN  & 52.5 \%		 \\ \hline
         IRLM (LCUs) & \textbf{60.8}\%  	\\ \hline
         RNN (384 block) & 55.0\%  	\\\hline
    \end{tabular}
    \subcaption{\label{tab:MRSC} Models initialized with long contexts}
    \end{subtable}    
    \caption{Results on text8 (a,b), Project Gutenberg (c) and the MRSC (c, d).}
    \vspace{-.25in}
\end{table}
\subsection{Skipping RNN extension}
\vspace{-.1in}
It is much easier to extend RNNs when gradient descent is used for learning, as opposed to when the Hessian-Free method is used. This can be important; as shown in \cite{Mikolov2012} character-level modelling generally lags behind word-level modelling in terms of predictive likelihoods so further work is needed to make character-level RNNs competitive. We present one enhancement we found with a simple extension of the RNN. We added skipping connections between the hidden units corresponding to the first letters of consecutive words. These connections formed a new matrix of parameters $\vec R_{\text{skip}}$ of the same size as $\vec R$. $\vec R_{\text{skip}}$ can potentially propagate information over longer distances in text. The skipping RNN is effectively a hybrid between a word-level and a character-level model. Adding in $\vec R_{\text{skip}}$ lowered the cross-entropy on the test set from 1.55 bpc to 1.48 bpc, while slightly increasing training times. This performance is still only about on par with state-of-the-art word-level n-gram models on this dataset, as estimated in \cite{Mikolov2012}. The skipping RNN may also extend to word-level language modelling. Instead of adding skipping connections between the first letters of words, we could add skipping connections between the first words of phrases, sentence clauses or even full sentences. We are currently investigating a hierarchy of such skipping connections but do not have any results yet. 

We note that a different solution for improving character-level language models has been discussed in \cite{Mikolov2012}, where the authors split words into their most frequent parts. These subparts consisted of short sequences of 1-3 characters. Indeed the HF-MRNN trained on the fragmented words achieved better cross-entropies that match the skipping RNN proposed here as can be seen in table \ref{tab:PennChar}. 
\vspace{-.2cm}
\section{Impulse response language model (IRLM)}
We considered completely giving up the nonlinearity $f$ and replacing it with the identity. Additionally, we set the recurrent matrix to be diagonal. We call this model an impulse response neural network. This model has completely linear propagation of the gradient in equation \ref{eq:BPTT}, for both positive and negative inputs. The IRLM computes a linear function $\vec x_t$ of the previous observations before passing $\vec y_{t+1}^T \vec Z \vec x_t$ through the softmax nonlinearity. Note however that the one-hot embedding of tokens is itself a highly nonlinear operation. The log-bilinear LM of \cite{MnihHinton2007} has a similar linear parametrization, however it lacks the long timescales in $\vec x_t$ which the IRLM can generate. As we will see in the results section, the IRLM is able to model the complex sequential patterns in language at the word-level. 

The linear contribution from $\vec y_{t-\tau}$ to $\text{P} \(\vec y_{t+1} | \vec y_t, \vec y_t,...\)$ before the soft-max nonlinearity can be rewritten as $\ds \vec y_{t+1}^T \vec Z \vec R^{\tau} \vec W \vec y_{t-\tau}$, regardless of whether $\vec R$ is diagonal or not. The LBL model of \cite{MnihHinton2007} uses a different parametrization where $\vec R^{\tau}$ is replaced with arbitrary interaction matrices $\vec D_{\tau}$. In fact, just like we do here, \cite{MnihHinton2007} use diagonal matrices $\vec D_{\tau}$. There are several advantages to using $\vec D_{\tau} = \vec R^{\tau}$. The first and most important is the implicit inductive bias the IRLM has for long timescales: words separated by several other words have an interaction term that only depends weakly on the length of the separation because $\vec R^{\tau}$ and $\vec  R^{\tau+1}$ are similar for entries of $\vec R$ close to 1. In the LBL model the interaction terms are entirely different at different delays $\tau$ and $\tau + 1$. Consequently the IRLM can generalize information learned at one delay $\tau$ to all other delays. For self-connections in $\vec R$ close to 1, the hidden units of the IRLM can be thought of as representing the topics of text. In fact, a recent state-of-the-art document topic model \cite{NADE} uses a similar parametrization of a neural network, there called NADE (neural autoregressor density estimation). NADE is the same model as the IRLM with self-connections of 1 but with an additional nonlinearity in front of $\vec x_t$ before multiplication with the matrix $\vec Z$. To model bag-of-words representations of documents, NADE assumes a random ordering of the words in each document. 

For optimization, we noticed that the gradient terms on the self-recurrent connections of the IRLM were very large compared to the gradients on the encoding and decoding matrices. This can be explained by the fact that the gradient of $R$ is updated on every iteration, whereas entries in the encoding matrix are only updated when their associated words are present. Entries in the decoding matrix, although non-zero on every iteration, are only large when their associated words are present due to the softmax nonlinearity. We used 1000-fold smaller learning rates for the self-recurrent connections than for the encoding and decoding parameters. We also considered a dynamic version of the IRLM, where the parameters of the model are dynamically adapted at test time in the direction of the gradient with respect to the new observations. We found modest but significant improvements, of a similar magnitude to those reported in \cite{Mikolov2011a}.
\vspace{-.2cm}
\section{Regularization with random dropout and column normalization}
\vspace{-.2cm}
We found that on small corpora like Penn, the IRLM was still able to overfit severely despite its relatively low-dimensional parametrization (compared to N-grams). On such problems, we applied the regularization method proposed by \cite{Hinton2012} and called there random dropout. The idea is to introduce noise into the function computed by the neural network so as to make the model more robust to test examples never seen before. The intuition provided by \cite{Hinton2012} is that the noise prevents units in the neural network from co-adapting their weights to overfit the training data. To avoid introducing instabilities into the recurrent part of the LMs, we added the noise only on the decoding portion of the model.

Formally we define $\eta$ to be a vector of Bernoulli random variables with probabilities 0.5 and length the dimensionality of the RNN. Then $\bar{\vec x}_t = \vec x_t \circ \eta$, and $\bar{\vec x}$ replaces $\vec x$ in equation \ref{eq:PYX} so that $\ds \text{P}_{\eta} \(\vec y_{t+1} | \eta, \vec y_t, \vec y_{t-1}, ...\) \propto \exp(\vec \vec y_{t+1}^T Z \bar{\vec x}_t)$. We define
\begin{align*}
\text{P} \(\vec y_{t+1} | \vec y_t, \vec y_{t-1}, ...\) = \langle  \text{P}_{\eta} \(\vec y_{t+1} | \eta, \vec y_t, \vec y_{t-1}, ...\) \rangle_{\eta}
\end{align*}
Because each $\ds \text{P}_{\eta}$ is a normalized probability, so is the average over all $\eta$-s. This is a slightly different generative model than the standard RNN, but it overfits less to the training data. A tractable lower bound on the likelihood of the noisy RNN model follows from a simple application of Jensen's inequality
\begin{align*}
\log \text{P} \(\vec y_{t+1} | \vec y_t, \vec y_{t-1}, ...\) &= \log \langle \text{P}_{\eta} \(\vec y_{t+1} | \eta, \vec y_t, \vec y_{t-1}, ...\) \rangle_{\eta}\\
&\geq \langle \log \langle \text{P}_{\eta} \(\vec y_{t+1} | \eta, \vec y_t, \vec y_{t-1}, ...\) \rangle_{\eta}
\end{align*}
We will estimate the gradient of this likelihood with samples from $\eta$. The resulting algorithm is exactly that of \cite{Hinton2012}. 

We also found that column normalization (CN) helped to further increase the performance of the IRLM on the Penn corpus. CN consists of fixing the L2 norm of the incoming/outgoing weights to each hidden unit. In the IRLM we find that low values of the column norm result in longer timescales. This is because large values of the hidden units can only be obtained by having large self-connections. In turn, large values of the hidden units are needed to generate low entropy predictive distributions for each word. Notice CN was also used in \cite{Hinton2012} in conjunction with the random dropout method but apparently for different reasons. The authors of \cite{Hinton2012} found that CN helped them use larger gradient steps during learning, while we found that CN helped generalization. Unfortunately the magnitude to which the columns of $\vec W$ and $\vec Z$ are normalized does influence the performance of the model so we had to crossvalidate this parameter separately by running 5 different experiments to find a good value (15) for the norm. One additional regularization strategy that we always used was to anneal the learning rates towards 0 on every epoch for which the validation cost decreased. 

\begin{figure}
\centering
     \includegraphics[width = 0.34 \textwidth]{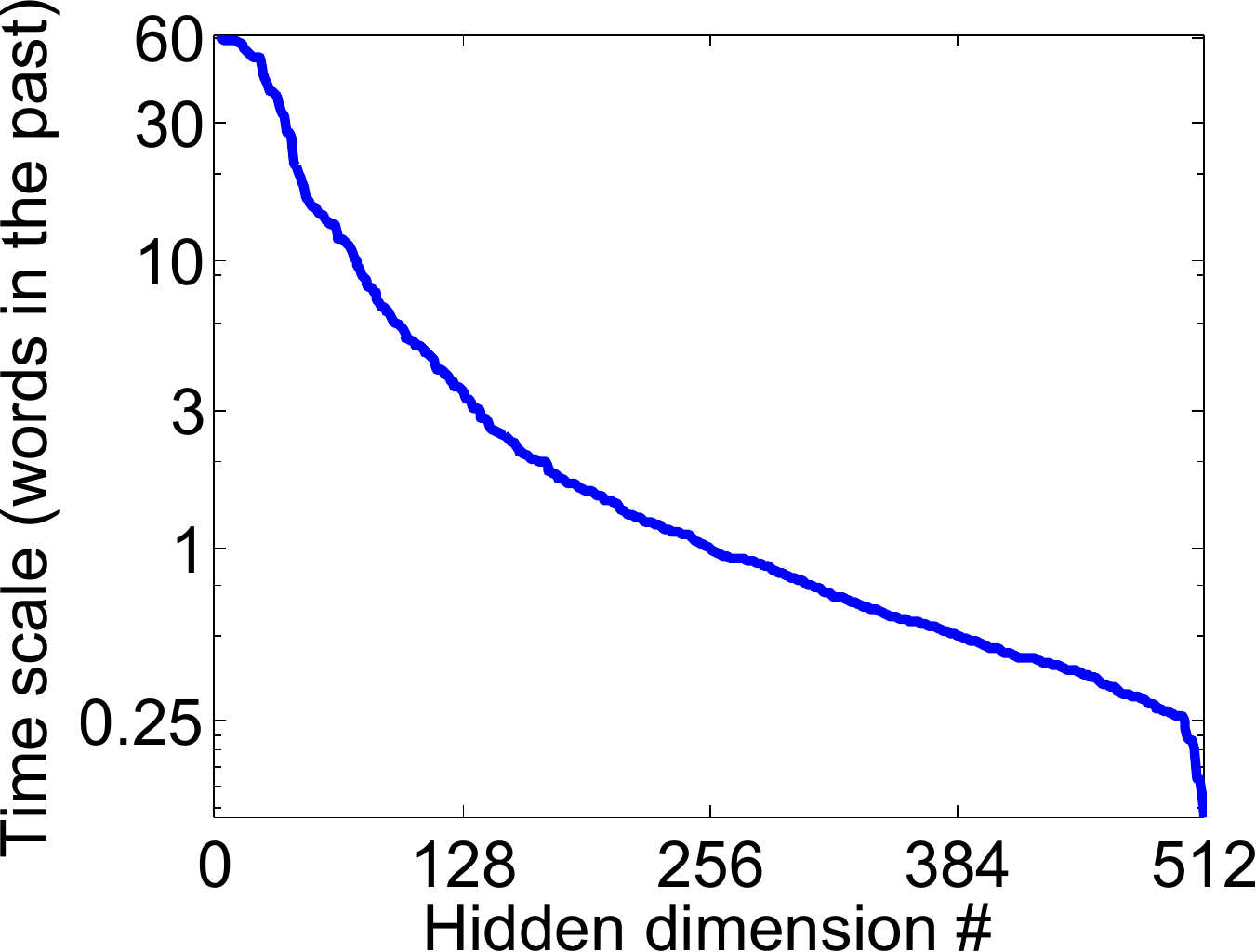}
     \centering
     \includegraphics[width = 0.34 \textwidth]{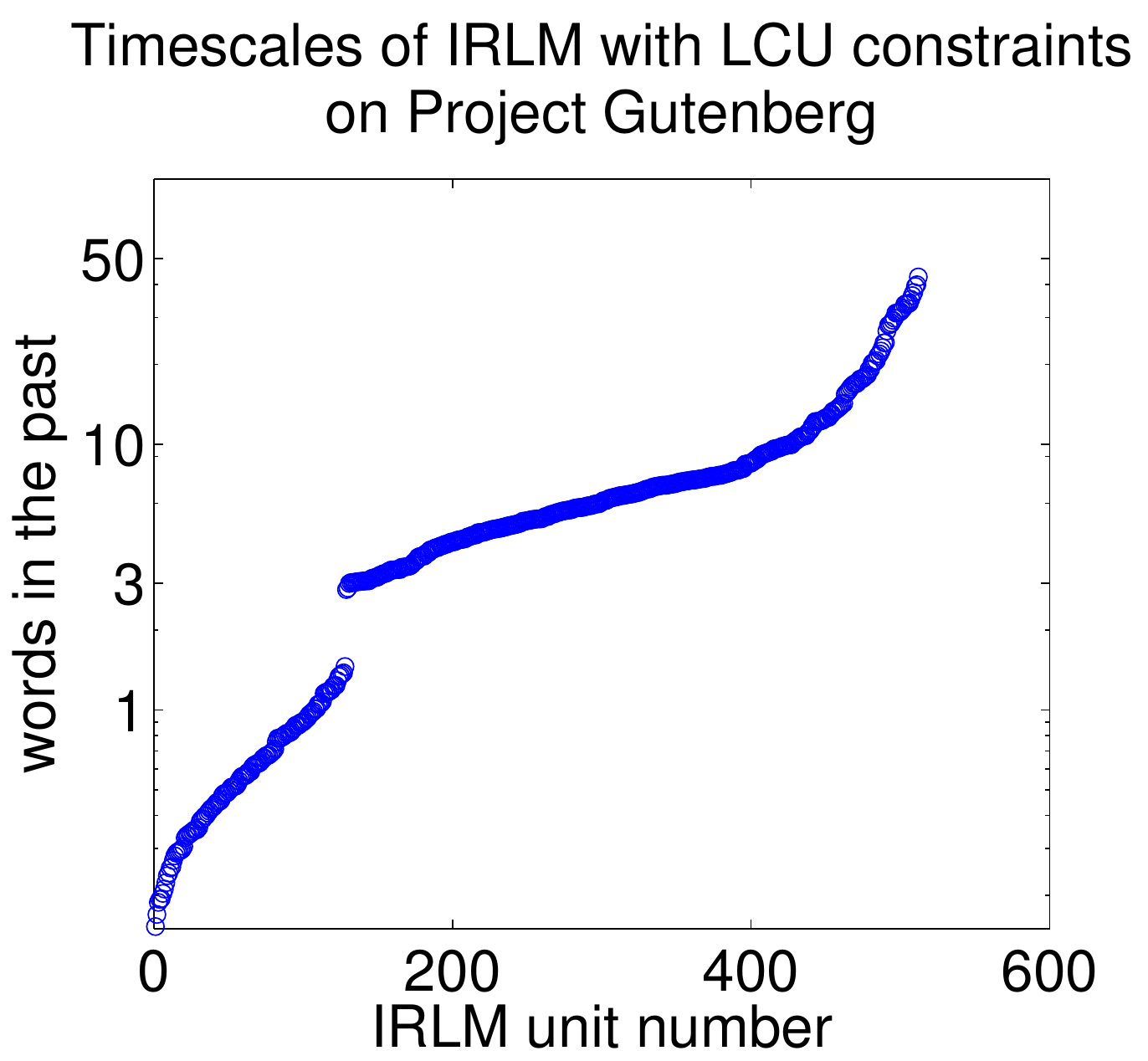}
    \caption{\label{fig:LCUs} a. The timescales of the IRLM on the Penn Corpus. b. The timescales of the IRLM initialized with LCUs on the Project Gutenber dataset. Note that the IRLM has discovered long-range dependencies in language, as indicated by the long timescales of up to fifty words. It is interesting that in our experiments with caching models, the unigram cache also only helps the perplexity up to a length of 50-100 words in the past.}
    \vspace{-.5cm}
\end{figure}
\vspace{-.2cm}
\section{Results}
\vspace{-.2cm}
We report the performance of the studied models on three datasets commonly used in the literature. The first is the Penn Corpus which contains 930k training word tokens. The second is the 'text8' dataset which contains a cleaned up version of the first 100 million characters from Wikipedia \footnote{\href{http://mattmahoney.net/dc/text8.zip}{http://mattmahoney.net/dc/text8.zip}}. Both of these corpora have well-defined cross-validation folds as described in \cite{Mikolov2012}, which allows us to directly compare our results to those reported elsewhere. Finally, we learned word-level models from the training set of the Microsoft Research Sentence Completion Challenge. The training dataset consists of about 500 copyright-free 19th century novels available through Project Gutenberg \cite{Zweig2011}. All of the models reported that we implemented have 512 hidden units, unless mentioned otherwise. This many hidden units are typical for word-level modelling. For the Penn dataset with a 10k vocabulary, this results in about 10 million parametes, while on the large datasets the models had about 70 million parameters. 
\vspace{-.2cm}
\subsection{Small dataset: Penn corpus}
It is commonly believed that the recurrent neural network (RNN) based models are able to capture highly nonlinear distributions of sequential data which simpler feedforward NNs cannot, thus explaining the better performance of RNNs for word-level language modelling. The unregularized IRLM achieves similar perplexities on the training data with the unregularized nonlinear RNN, meaning they can capture similarly many patterns (table \ref{tab:PennWord}). However, the RNN generalizes better and achieves lower perplexities on test data. This means the nonlinear nature of the RNN serves as a kind of regularizer, allowing the network to learn some important patterns present in language and preventing some spurious associations to be made, which the feedforward models make more easily. As a generative model, the RNN can be said to have better inductive biases than a feedforward neural network, perhaps owing to its dynamical representation and nonlinearities. However, after random dropout regularization, we find that the performance of the two models considered here increases significantly and to similar levels (table table \ref{tab:PennWord}). The regularized IRLM still overfits slightly more to the training data than the regularized RNN, but their predictive performances are nearly the same. The IRLM but not the RNN is further improved by column normalization. The regularized IRLM achieves the state-of-the-art result on the Penn corpus for a single model with a perplexity of 102.5 down from the previous best published result of 123 \cite{Mikolov2011a}. Mixtures of models usually fare better and can enhance performance to a perplexity of 77 when a large number of very different models are averaged together \cite{Mikolov2011a}.

In the literature, it is common to also report the results of models averaged with KN5, unigram caches and/or averaged with multiple copies of the model trained independently. All of these further improve the results considerably as shown in table \ref{tab:PennWord}. We highlight that of special interest are results for single models, or for single models combined with n-gram based models, as these will have the lowest computational cost at runtime and lowest memory trace. This cost can be significant with large vocabularies. Also, the Penn dataset is very small and on larger datasets training more than one model can be computationally expensive and memory taxing.
\begin{table}
    \footnotesize
    \centering
    \begin{tabular}{ | l | c | c | c | c | c |}
    \hline
     & Single train/test& +KN5+cache & x10 & x10+KN5+cache\\ \hline
     5-gram Kneser-Ney$^1$ & 10/141.2 & 125.7 &&\\ \hline
     feedforward NNLM$^1$ & ?/140.2 & 106.6 &&\\ \hline
     Log-bilinear LM$^1$ & ?/144.5 & 105.8 &&\\ \hline

    RNN$^1$  & 124.7 & 97.5 & 102.1 & 89.4\\ \hline
    dynamic RNN$^1$  & ?/123.2 & 98.0 & 101.0 & 90.0\\ \hline
    &&&&\\ \hline
    RNN$^3$ (no reg) & 34/126 & & & \\ \hline
    RNN$^3$ ($^2$DO\&CN) & 40/\textbf{107} & & & \\ \hline
	 &&&&\\ \hline
    IRLM(no reg) & 32/137 &   &&\\ \hline
    IRLM(L1 reg) & ?/125 &   &&\\ \hline
    IRLM ($^2$DO) & 38/\textbf{109} & &&\\ \hline
    IRLM ($^2$DO\&CN) & 42/\textbf{102.5} & \textbf{94}& 98.5 & 92.5\\ \hline
    dynamic IRLM($^2$DO\&CN) & ?/\textbf{98.5} & \textbf{90.5}  & 95 & 89\\ \hline
    \end{tabular}
    \caption{\label{tab:PennWord} Results on Penn corpus. \\$^1$ as reported in \cite{Mikolov2011a},
        $^2$ trained with random dropouts and column normalization, $^3$ our implementation
}
\end{table}
One interesting aspect of the IRLM is that we can directly assess how large of a context the network uses by looking at the diagonal matrix $\vec R$. The effective contribution to $\vec x_t$ from a timelag $\tau > 0$ in the past is $\ds \vec R^{\tau-1} \vec W \vec y^{t-\tau}$ which then decays like $\ds \vec R^{\tau} = \exp(\log(\vec R) \tau)$. The timescales of the network are defined as $\ds -1/\log(R)$, where the division operation is understood elementwise on the diagonal of $\vec R$. For an IRLM trained on the Penn corpus we plot these timescales in figure figure \ref{fig:LCUs}a. One can see that the IRLM has learned several very long timescales of up to 50 words. This reminds of the benefits offered by caching methods to language models: many language models are improved at test time if the predictive probability of the model is interpolated with a unigram model learned exclusively from the previous 50-100 words. We can see that nonetheless most of the timescales of the IRLM are relatively short in order to model the local grammar of language. Each experiment on the Penn dataset took about three hours to run on a GTX 680 GPU.

\subsection{Large datasets: 'text8' and Project Gutenberg}
We also ran the same experiments on a much larger corpus known as 'text8', consisting of the first 100M characters of Wikipedia. As a preprocessing step we considered only words that appear at least five times in the training set, converting all others to the special '$<\text{unk}>$' token. This left us with a large vocabulary of 67428 words and a training corpus of 15301600 words. Training RNN models on this much data would require several days for a single experiment so we turned to the fast approximate training method called noise contrastive estimation (NCE), proposed by \cite{GuttHyv2010} and adapted for neural language modelling by \cite{MnihTeh2012}. The reader is advised to read \cite{MnihTeh2012} for full details of the procedure. Briefly, the method involves training a classifier to distinguish between real sampled sequences of words and sequences where one word is sampled from noise distributions, where the classifier is based on the generative model being trained. The perplexity results that we get with NCE trained neural language models are significantly better than those of n-gram models. In addition, the RNN-LM achieves 10\% lower perplexity than IRLM on both training and test data, indicating that the RNN has a larger capacity for storing sequences \footnote{In the first version of this arXiv paper we reported a perplexity of 191 for the RNN model on this dataset. That result was after 10 epochs of training, which is usually considered sufficient especially for such large datasets. However, in further experiments we found that RNN but not IRLM models improved up until 50 epochs of training to a perplexity of 179, resulting in significant perplexity differences between the two models. We thank Yoshua Bengio for suggesting that this might be the case.}. 

We also ran experiments on the training dataset for the Microsoft Sentence Completion Challenge \cite{Zweig2011}. This dataset consists of about 500 19th century novels from the Project Gutenberg database. The RNN-LM was again able to capture many more patterns in the data compared to the IRLM, resulting in a 20\% perplexity difference. 

We find that dropout regularization no longer helps generalization with this large amount of training data. Instead, the performance of the regularize and non-regularized moedls is almost the same, and much better ($\sim$ 35\%) than the vanilla n-gram model to which comparisons are usually made in the literature (Kneser-Ney of order 5). However, cache models significantly improve n-grams and in this case by a very large margin of $\sim$ 25\% perplexity. The neural language models offered improvements of 20-30\% perplexity over that. 

\subsection{Microsoft Research Sentence Completion Challenge}
The MRSC challenge consists of 1040 SAT-style sentences with blanked words and five possible choices for each blank. The task is to choose the sentence that makes most sense out of the five possible ones. Humans perform on this task at around $\sim$ 90 \% correct \cite{Zweig2011}. The training set for language models on this task consists of the Project Gutenberg novels. N-gram models perform poorly in the MRSC challenge at around 40 \% correct but RNN-LMs perform at around 50\%, while the LBL model of \cite{MnihTeh2012} obtains 54.7\%. Although \cite{Mikolovthesis} reports results for RNN-LMs, it might be that the advantage of the LBL model is related to the more efficient training method, thus resulting in more epochs of learning. We trained both RNN-LMs and IRLMs on the Project Gutenberg novels using NCE, and report in table \ref{tab:perpGUT} their respective performance. 

In agreement with \cite{Mikolovthesis} we find that the NCE-trained RNN-LM performs at $\sim$ 50\% correct. Despite having a much lower perplexity, the IRLM performs slightly better at 52.5\%, though not as well as the LBL model reported in \cite{MnihTeh2012}. This ordering is highly surprising, because a 20\% perplexity gap exists between the RNN-LM and the IRLM. We conjecture that due to its restricted representation the IRLM needs to focus more than the RNN on the long range dependencies in language, which help it have better semantic comprehension. To verify our conjecture, we ran an IRLM experiment where all the units were initialized to 0.9. These units effectively `see' about ten words into the past and we call these long context units (LCUs). During training, a proportion of about 25\% of the units dropped in value to code for short context dependencies but the other units remained above 0.5. This model scored 0.55\% correct on the MRSC task. We then designed another IRLM in which we enforced LCUs to keep large values by always constraining them to lie between 0.7 and 1. We used 128 short-context units initialized at 0 and 384 LCUs. After training, we used only the values of the LCUs to make predictions about which MRSC sentence is correct by setting the values of the short-context units to 0. This is potentially highly disruptive for the normalization constant in each context, so we did not normalize the predictions of the LCUs. The LCUs alone achieved 60.8\% correct which is a new state-of-the-art result on this task. After learning, most of the LCUs were in the range of 0.7 to 0.9 (figure \ref{fig:LCUs}b), suggesting the relevant timescales analyzed are between 3 and 10 words in the past.  

We attempted similar experiments with RNN-LMs. To this end we partitioned the recurrent matrix into two blocks of units and set connections  between blocks to 0. We initialized connections within the larger block of 384 units with 0.9 on the diagonal and the smaller block of 128 units with all zeros. This way, like the LCUs, the 384 block of the RNN is biased by initialization to keep information from long contexts. However, it was not possible to enforce this constraint during learning. The best result we could get using the 384 unit block of the RNN was 55\% correct, and only when using very small learning rates for the 384 unit block. Initializing the 384 unit block to an Echo State Network as proposed in \cite{Sutskever2013} was even less succesful. 

\section{Conclusions}
We have presented language modelling experiments with RNNs and IRLMs aimed at evaluating the two models' intrinsic regularization properties, storage capacity and ability to capture long contexts. On a small dataset we found that the regularization method (random dropout) was far more important than the model used. The best model was an IRLM that scored 102.5 perplexity on the test set, and used several regularization techniques: random dropout, column normalization and annealed learning rates. On large datasets, the high capacity of the RNN allows it to store and recognize more patterns in language. Nonetheless, on a sentence comprehension challenge that requires integrating information over long contexts we found that the IRLM was slightly superior. In addition, the IRLM's simple representation allowed us to use only the long context units for scoring sentences, which resulted in a large boost in accuracy to 60.8\% correct. This represents a new best result on this dataset improving on the 54.7\% from \cite{MnihTeh2012}. While the same information from long contexts might in principle be embedded in the representation of the RNN, it is not straightforward to obtain. We see that there is reason to develop more easily accessible representations for neural language models and IRLM constitues a first step in that direction.

\section*{Acknowledgements}
  We thank Andriy Mnih and Yoshua Bengio for reading versions of this article and providing helpful comments. We also thank Andriy for introducing us to neural language models and for suggesting the name IRLM and 	we thank Yoshua for pointing out that the nonlinear dynamics of the RNN really should matter for storing patterns and making predictions. 

\bibliographystyle{unsrt}
{\small
\bibliography{langbib}}

\end{document}